Crowd collectiveness measure via graph-based node clique learning

Ren weiya[1,2]

Email: weiyren.phd@gmail.com; renweiya@nudt.edu.cn.

(1.Department of Management Science and Engineering, Officers College of Chinese Armed Police Force, Sichuan Chengdu 610213, China.

2.College of Information System and Management, National University of Defense Technology, Hunan Changsha, 410073, China)

**Abstract:** Collectiveness motions of crowd systems have attracted a great deal of attentions in recently years. In this paper, we try to measure the collectiveness of a crowd system by the proposed node clique learning method. The proposed method is a graph based method, and investigates the influence from one node to other nodes. A node is represented by a set of nodes which named a clique, which is obtained by spreading information from this node to other nodes in graph. Then only nodes with sufficient information are selected as the clique of this node. The motion coherence between two nodes is defined by node cliques comparing. The collectiveness of a node and the collectiveness of the crowd system are defined by the nodes coherence. Self-driven particle (SDP) model and the crowd motion database are used to test the ability of the proposed method in measuring collectiveness.

# 1 Introduction

In crowd surveillance, the collectiveness motions of crowd have attracted a great deal of attentions in recent years [2]-[6],[7][14][15]. Due to collectiveness motions are universe, intuitionistic and macroscopic, measuring the collectiveness motions across different scenes is meaningful for crowd behaviors comprehension [11][19].

The path integral descriptor method [11] is a novel method to quantify the structural properties of collective manifolds of crowds across different scenes. Path integral [1], which was first introduced in statistical mechanics and quantum mechanics [1][8][9][10][12], sums up the contributions of all possible paths to the evolution of a dynamical system and is free from the restriction on data distributions [16]. [1] uses a generating function to produce the path integral descriptor, which is adopted in [11]. Based on [1] and [11], [19] uses exponent generating function to sum the path integral and compute the path integral descriptor to avoid parameter setting. The path integral descriptor is the graph based method, which use a graph with nodes to represent the whole crowd systems. The path integral descriptor is actually the measurement of collectiveness, and large value of the path integral descriptor means that nodes in the set have a high coherence motion.

The path integral descriptor methods study the coherence among nodes by path integral descriptors, and investigate all possible paths of a pair of nodes. All paths of a pair of nodes may be infinite for path length varies from zero to infinite, which is an ideal math model. Different form path integral descriptor methods, we propose a node representation learning method in this paper, which named Node clique learning (NCL) method. The proposed method is also a graph based method and investigates the influence from one node to other nodes. Generally speaking, we represent a node by a set of nodes which named a clique.

To compute the clique of a node, we first endow this node with certain information and spread the information to other nodes. Due to the weight on edge ranges from 0 to 1, information will not increase.

If the information decreases to a low value, we end the spreading process and collect nodes as the clique. It means only nodes with enough information are collected as the clique. In information spreading step, we use two strategies which are the average strategy and the min strategy to learn an unrenewed node's information. After clique learning, the coherence among nodes can be computed by cliques comparing. In clique comparing step, we use two ways to define the coherence among nodes. At last, the collectiveness of a node can be defined as the averaging coherence between this node and all other nodes, and the collectiveness of the crowd system can be computed as the average collectiveness of nodes.

We compare the proposed method with the state-of-the-art method on self-driven particle (SDP) model [13] and crowd motion database[11]. The paper is organized as follows. The proposed node clique learning method is proposed in Section 2. Then, the collectiveness measure experiment is presented in Section 3. Conclusion is given in Section 4.

# 2 Node clique learning

In this section, we first introduce the neighborhood graph of a system. Then we show how to learn the clique of each node in the graph. After obtaining cliques of nodes, we compare the cliques of each pairs of nodes to define their motion coherence. At last, the motion coherence is used to define the collectiveness of a system.

## 2.1 Neighborhood graph

Given a set $\mathcal{C}$ of samples $X = [x_1, x_2, …, x_n] \in R^{D \times N}$, which $D$ is the dimension of data and $N = |\mathcal{C}|$ is the number of samples. Then, we build a directed graph $G = (V, E)$, where $V$ is the set of nodes (vertices) corresponding to the samples in $X$, and $E$ is the set of edges connecting nodes. The graph is associated with a weighted adjacency matrix $W$, where $w_{ij} \in [0,1]$ is the weight of the edge from vertex $i$ to vertex $j$. The famous $K$-NN graph is adopted as the edge connection strategy, which means each node has $K$ edges pointing from itself to its $K$ nearest neighbors. Due to each node has $K$ outcome edges, $K$-NN graph permits edges with zero weights.

## 2.2 The clique of a node

When the graph $G = (V, E)$ is given, we first need to learn the clique of each node in the graph. The clique of one node is actually a set of nodes, which is obtained by spreading information from this node to other nodes in the graph. In the process of learning the clique of one node, we can name this node as core node and name the others as normal nodes. Note that the core node becomes to a normal node when learning the cliques of other nodes, while a normal node becomes to the core node when learning its own clique.

Before the information spreading process, we assume only the core node has the information and we set its information to one. At the same time, the normal nodes have no information and their information value can be set as zero. When the information process begins, the core node spreads its information to the normal nodes. After the information spreading, some nodes have enough information and some others have insufficient information or no information. Enough information means that the information is larger than a predefined threshold. We collect nodes with enough information as the clique of the core node. In this way, we can represent a node by a clique, which is a set of nodes.

The information spreading process is the key in node clique learning. We now show the details of the information spreading. At the beginning, only the core node is renewed and all normal nodes are unrenewed. A normal node is regard as renewed only when its information is computed. To compute the information of an unrenewed normal node, we can consider its neighbors. We hold that only nodes with enough information have relations with the core node. To be concise, these nodes can be entitled as privileged nodes. Thus, we can compute the information of an unrenewed normal node by its privileged neighbors. We also call a node as "ready node" if this node is unrenewed and has at least one privileged node. In this way, we can achieve the information process by iteratively computing the ready node's information.

In general, we compute a clique of a given node as follows:
1. Give a graph and a node. This node is regard as core node and the other nodes are regard as the normal nodes. We set the core node's information to one, and make its state as renewed and privileged. At the same time, all the normal nodes are unrenewed and not privileged.
2. Find a ready node. A ready node is a normal node which is unrenewed and has at least one privileged neighbor. We compute the ready node's information by its privileged neighbors. The edge weight from the privileged node to the ready node can be seen as the weight of the privileged node. Two strategies are considered: ①Average strategy. The average of the weighted privileged neighbors. ② Min strategy. The min value of the weighted privileged neighbors.
3. An unrenewed node becomes renewed when its information is computed. After computing one node's information, we give it a privilege only when its information is larger than a predefined threshold.
4. The information spreading process ends only when there are no ready nodes any more in the graph. Otherwise go back to step 2.
5. When the information spreading process is over, we collect all privileged nodes as the clique of the core node.

In step 2, there are two strategies to compute the ready node's information. The Average strategy emphasizes the general information using of the privileged neighbors. The Min strategy takes a cautious approach by adopting the min value of the weighted privileged neighbors.

An example of learning the clique of node is illustrated in Fig.1 and Fig. 2. Fig.1 is the illustration of the drawing. In Fig.2, node 1 is the core node with information value 1 and its state is renewed and privileged. The information threshold is 0.6 in this case. At first iteration, node 2 is one of the ready nodes and its information is computed by its privileged neighbors, which are node 1. At second iteration, node 3 is one of the ready nodes and its information is computed by its privileged neighbors, which are node 1 and node 2. The computed information of node 3 is not larger than 0.6, thus node 3 is not privileged. The information spreading process ends when there are no ready nodes any more. Finally, the clique of node 1 is {1,2,4,5}.

## 2.3 Clique comparing

We compute all nodes' clique by node clique learning in section 2.2. In this way, each node is represented by a clique. Then the cliques are used to compute the coherence (or similarity) among nodes. The coherence of two nodes can be computed by comparing their cliques. To compare two nodes' coherence, it is reasonable to consider the following two criterions:
1. The similarity of two nodes' cliques.
2. The frequency of mutually occurring in other nodes' cliques.

If two nodes have high coherence, their cliques would be similar. Besides, they would have a high frequency of mutually occurring in other nodes' cliques, which also means they do not occur alone in other nodes' cliques in most cases.

The clique is actually a set of nodes. **Jaccard similarity coefficient** can be used to measure the similarity of two sets. Give two sets $A$ and $B$, the Jaccard similarity coefficient of $A$ and $B$ is defined as

$$J(A,B) = \frac{|A \cap B|}{|A \cup B|}. \tag{1}$$

To be concise, we first represent the cliques by a matrix. The cliques of nodes is denoted as a matrix $C \in R^{N \times N}$. Each row of $C$ represents a clique of node. The $i$-th $(i = 1,2,...,N)$ row of $C$ is denoted as $C_i \in R^{1 \times N}$. Thus, $C_i$ is the clique representation of node $i$. $C_i(j)$ $(j = 1,2,...,N)$ is the $j$-th element of $C_i$ and $C_i$ is defined as

$$C_i(j) = \begin{cases} 1, & \text{node } j \text{ belongs to the clique of node } i. \\ 0, & \text{otherwise.} \end{cases} \tag{2}$$

In this way, the similarity between the clique of node $i$ and the clique of node $j$ is computed as

$$J_1(i,j) = \frac{\sum(C_i \circ C_j)}{\sum max(C_i, C_j)}. \tag{3}$$

where $\circ$ denotes element-wise product of vectors, $max(\cdot,\cdot)$ denotes element-wise max comparison of vectors.

We denote the $i$-th $(i = 1,2,...,N)$ column of $C$ as $C^i \in R^{N \times 1}$. The frequency of two nodes mutually occurring in other nodes' cliques can be compute as

$$J_2(i,j) = \frac{\sum(C^i \circ C^j)}{\sum max(C^i, C^j)}. \tag{4}$$

The coherence of node $i$ and node $j$ can be defined as

$$Z(i,j) = \begin{cases} \dfrac{[J_1(i,j) + J_2(i,j)]}{2}. & i \neq j \\ 0. & i = j \end{cases} \tag{5}$$

At last, we can obtain a matrix $Z \in R^{N \times N}$ to represent the coherence among nodes.

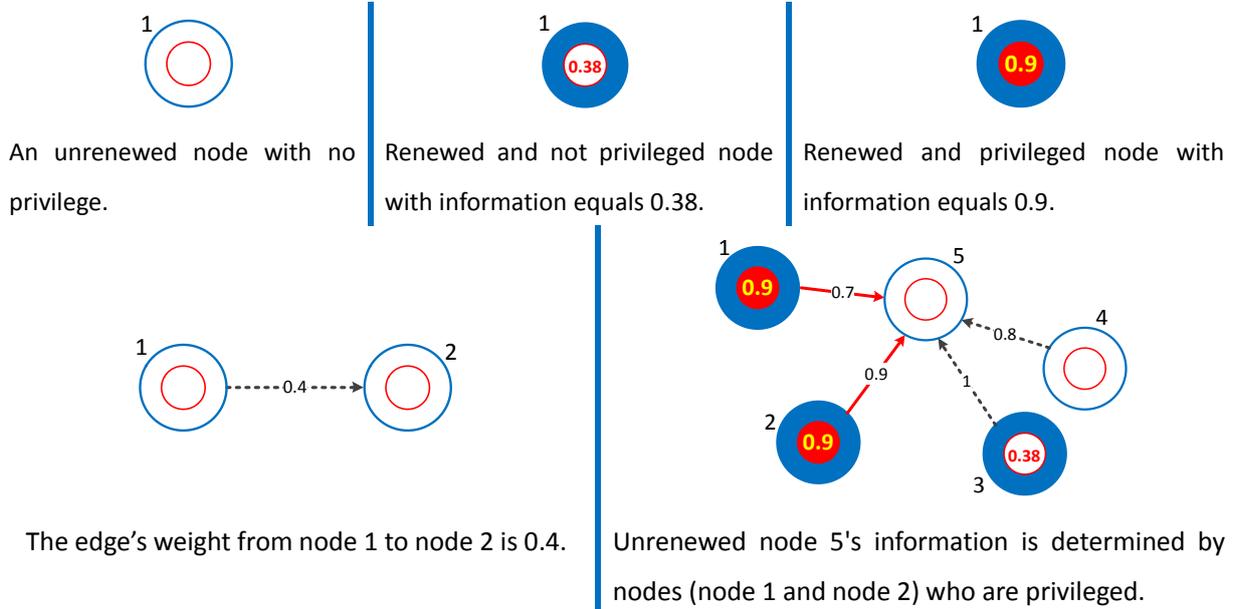

Fig. 1. Illustration of the drawing.

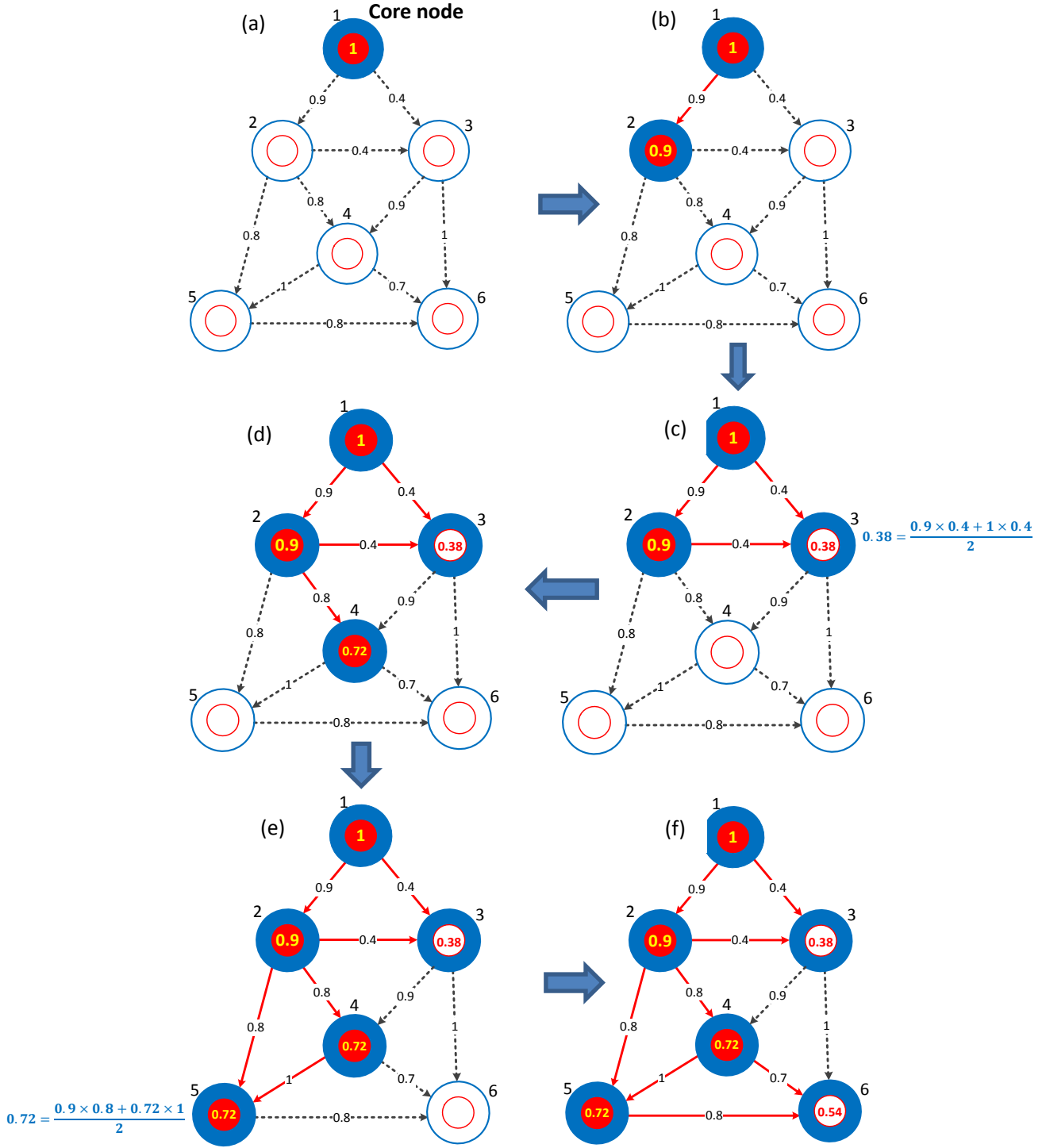

Fig. 2. An example of learning the clique of node 1 by NCL1-avg, and the clique of node 1 is {1,2,4,5}.

If we consider the asymmetry similarity of clique nodes, the **Absolute similarity coefficient** of sets can be adopted. The similarity of set $B$ to $A$ can be define as

$$J'(A,B) = \frac{|A \cap B|}{|A|}. \tag{6}$$

In this way, the similarity between the clique of node $i$ and the clique of node $j$ is computed as

$$J'_1(i,j) = \frac{\sum(C_i \circ C_j)}{\sum C_i}. \tag{7}$$

The frequency of two nodes mutually occurring in other nodes' cliques is compute as

$$J'_2(i,j) = \frac{\sum(C^i \circ C^j)}{\sum C^i}. \tag{8}$$

The coherence of node $i$ and node $j$ can also be defined as

$$Z(i,j) = \begin{cases} \frac{[J'_1(i,j) + J'_2(i,j)]}{2}. & i \neq j \\ 0. & i = j \end{cases} \tag{9}$$

## 2.4 Collectiveness measure

### 2.4.1 Node collectiveness

If nodes in a set have a high coherence with each other, the collectiveness of a graph should be high. Thus, the coherence of node cliques can be comprehended as the measurement of collectiveness of the set.

If a node has high collectiveness in the graph, it would have high coherence with other nodes. The collectiveness of a node can be defined as the averaging coherence between this node and all other nodes. We denote the collectiveness of node $i$ $(i = 1,2,...,N)$ as $\varphi(i)$, then $\varphi(i)$ can be computed as

$$\varphi(i) = \frac{\sum_{j=1}^{N} Z(i,j)}{N-1}. \tag{10}$$

where $Z(i,i) = 0$.

Node clique is an important measuring index of nodes. If a node has high coherence with other nodes, it means it has high node collectiveness. If a node has a high collectiveness, it means it plays an important role in the graph. Fig 3 shows an example of node clique learning and node collectiveness computing. We can observe that the node has high coherence with other nodes will have high node collectiveness. High collectiveness also means high influence in the graph.

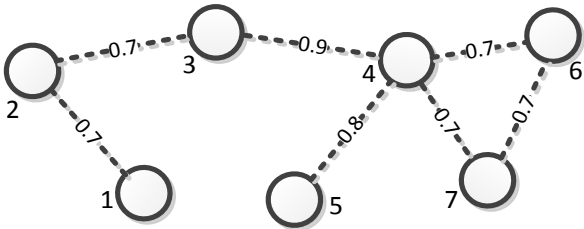

(a) A undirected graph.

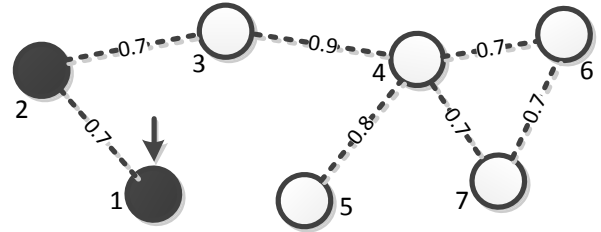

(b) The clique of node 1 (marked in black).

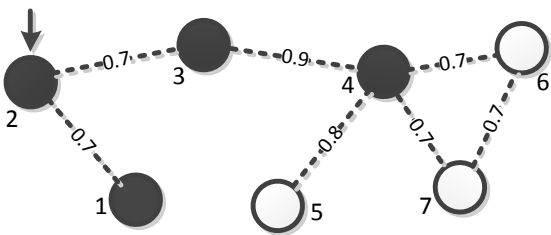

(c) The clique of node 2.

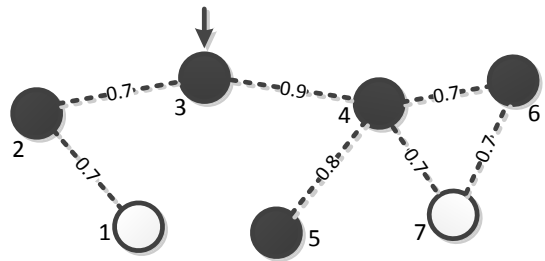

(d) The clique of node 3.

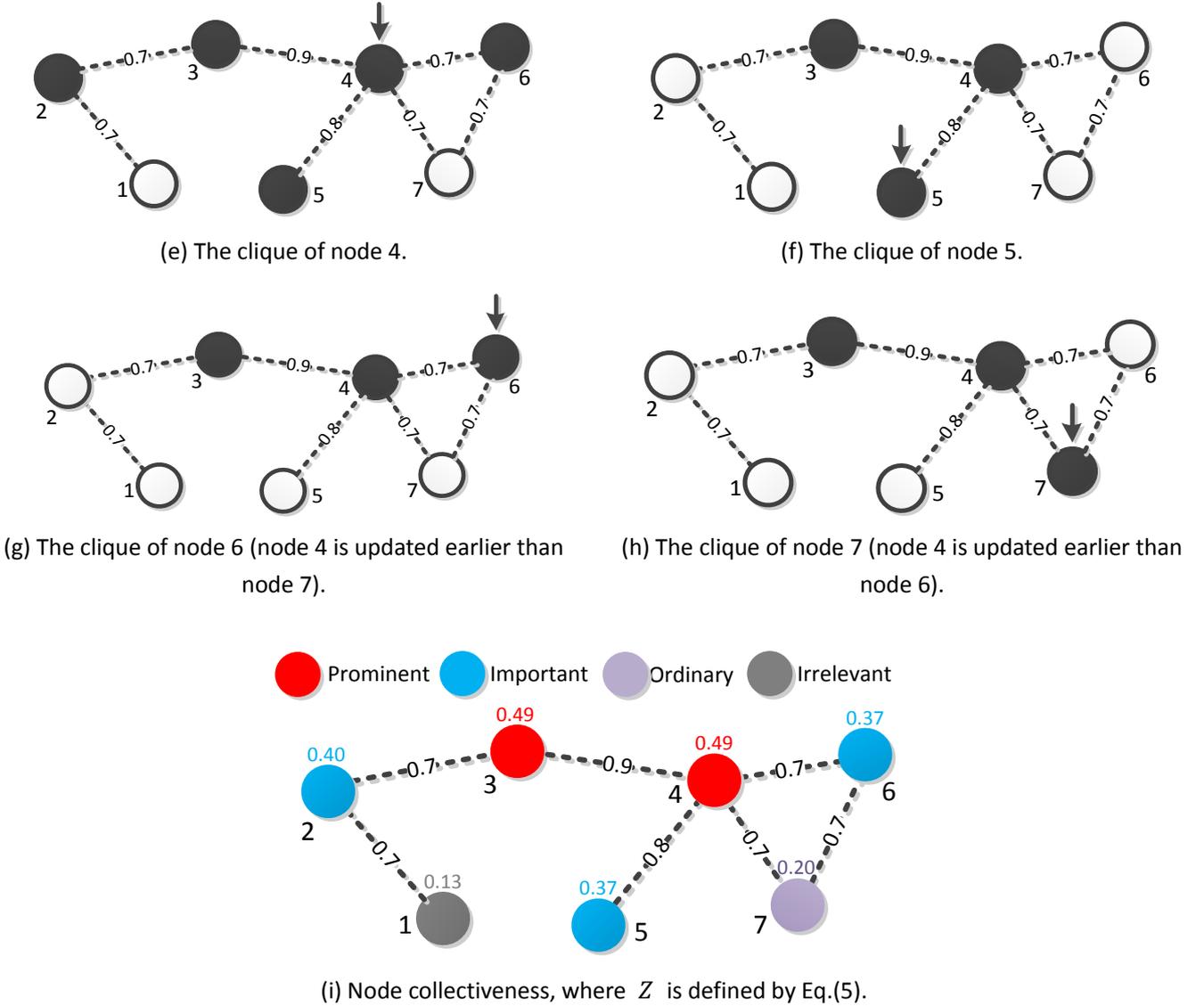

(e) The clique of node 4.

(f) The clique of node 5.

(g) The clique of node 6 (node 4 is updated earlier than node 7).

(h) The clique of node 7 (node 4 is updated earlier than node 6).

(i) Node collectiveness, where $Z$ is defined by Eq.(5).

Fig. 3. An undirected graph is given in (a). Cliques of nodes are shown in (b)~(h). Collectiveness of nodes is shown in (i).

### 2.4.2 Graph collectiveness

The collectiveness of the graph can be computed as the average collectiveness of nodes. We denote the collectiveness of graph as $\Phi$ then $\Phi$ can be computed as

$$\Phi = \frac{\sum_{i=1}^{N} \varphi(i)}{N}. \tag{11}$$

## 2.5 Algorithm details

The whole algorithm of the proposed Node Clique Learning method (NCL) is described in algorithm 1. In addition, NCL is named as NCL1 and NLC2 if $Z$ is defined by Eq.(5) and Eq.(9), respectively. There are two strategies (underline part in algorithm 1) to compute the ready node's information. We name NCL as NCL_avg (NCL1_avg and NCL2_avg) and (NCL1_min and NCL2_min) if we adopt the average strategy and the min strategy, respectively.

**Algorithm.1 Collectiveness measure algorithm via Node Clique Learning (NCL) method**

**Input**: The $K$-NN graph $W$ ($K = 20$) of dataset $X$, and the parameter $0 \leq \lambda \leq 1$.
**Initialization**: $\lambda = 0.7$.
**Denotes:** ①The information of node $i$ is denoted as $V_i$.
②We denote $P_i = 1$ if node $i$ is privileged and denote $P_i = 0$ otherwise.
③We denote $R_i = 1$ if node $i$ is renewed and denote $R_i = 0$ otherwise.
④The neighbors of node $i$ is denoted as $Y(i)$.
⑤The privileged neighbors of node $i$ is denoted as $\Theta(z)$, $\Theta(z) \triangleq \{x | x \in Y(z) \text{ and } P_x = 1\}$. We denote the number of elements in $\Theta(z)$ as $n_z$.
⑥The clique of node $i$ is denoted as $Clique_i$.
⑦The ready nodes set is denoted as $C_t$ at $t$-th iteration. The $z$-th node in $C_t$ is denoted as $C_t(z)$. The number of nodes in $C_t$ is denoted as $n_t$.
**For node** i $(i = 1,2, ..., N)$
　Initializations:
　　①Information: $V_i = 1$, $V_j = 0$ $(j = 1,2, ..., N, j \neq i)$.
　　②Priviledged state: $P_i = 1$, $P_j = 0$ $(j = 1,2, ..., N, j \neq i)$.
　　③Renewed state: $R_i = 1$, $R_j = 0$ $(j = 1,2, ..., N, j \neq i)$.
　　④The iteration: $t = 1$.
　Find $C_t$: Node $z \in C_t$ only if $R_z = 0$ and $\sum_x P_{x \in Y(z)} > 0$ ($z$ has at least one privileged neighbor).
　**While** $C_t$ **is not empty**
　　**For node** $z$ $(z = C_t(1), C_t(2), ..., C_t(n_t))$
　　　①$R_z = 1$.
　　　②Find $\Theta(z)$.
　　　③<u>Average strategy</u>: $V_z = (\sum_x w_{xz} V_x)/n_z$, where $x \in \Theta(z)$. **Or** <u>Min strategy</u>: $V_z = min(w_{xz} V_x)$, where $x \in \Theta(z)$.
　　　④$P_z = 1$ only if $V_z > \lambda$.
　　**End for**
　　$t = t + 1$.
　　Find $C_t$.
　**End while**
　$Clique_i = \{x | V_x > \lambda\}$.
**End for**
Obtain the cliques matrix $C$ by Eq. (2).
Compute the node coherence matrix $Z$ by Eq.(5) **Or** Eq.(9).
Compute the collectiveness of nodes ($\varphi$) by Eq.(10).
Compute the collectiveness of the graph ($\Phi$) by Eq.(11).
**Output**: $(C, Z, \varphi, \Phi)$.

## 2.6 Properties and parameters of NCL

### 2.6.1 Properties

**Property 1. (Convergence and range).** $Z$ *always converges and* $0 \leq Z(i,j) \leq 1$.

**Proof.** Since $0 \leq \frac{\Sigma(C_i \circ C_j)}{\Sigma max(C_i, C_j)} \leq 1$ and $0 \leq \frac{\Sigma(C^i \circ C^j)}{\Sigma max(C^i, C^j)} \leq 1$, we have $0 \leq (\frac{\Sigma(C_i \circ C_j)}{\Sigma max(C_i, C_j)} + \frac{\Sigma(C^i \circ C^j)}{\Sigma max(C^i, C^j)})/2 \leq 1$. Since $0 \leq \frac{\Sigma(C_i \circ C_j)}{\Sigma C_i} \leq 1$ and $0 \leq \frac{\Sigma(C^i \circ C^j)}{\Sigma C^i} \leq 1$, we have $0 \leq (\frac{\Sigma(C_i \circ C_j)}{\Sigma C_i} + \frac{\Sigma(C^i \circ C^j)}{\Sigma C^i})/2 \leq 1$. Thus, we know $0 \leq Z(i,j) \leq 1$ by both Eq.(5) and Eq.(9). Thus, $Z$ always converges. ∎

**Property 2.** $J_1 = J_2$, $J'_1 = J'_2$ *when* $W$ *is a symmetry matrix*.
**Proof.** If $W$ is a symmetry matrix, then edges in graph are undirected. From the defining of Eq. (2), we know $C$ is a symmetry matrix. Then we have $C^i = C_i$. Thus, $J_1 = J_2$, $J'_1 = J'_2$. ∎

**Property 3.** $Z$ is a symmetry matrix when defined by Eq.(5).
**Proof.** Due to Eq. (3), $J_1(i,j) = J_1(j,i)$. Thus, $J_1$ is a symmetry matrix. In the same way, $J_2$ is also a symmetry matrix. Due to Eq. (5), we know $Z$ is a symmetry matrix. ∎

**Property 4. (Bounds of $\Phi$).** $0 \leq \Phi \leq 1$.

**Proof.** Since $0 \leq Z(i,j) \leq 1$ and $Z(i,i) = 0$, we have $0 \leq \frac{\sum_{j=1}^{N} Z(i,j)}{N-1} \leq 1$ $(i = 1,2,\ldots,N)$. It means $0 \leq \varphi(i) \leq 1$. Then we have $0 \leq \frac{\sum_{i=1}^{N} \varphi(i)}{N} \leq 1$, which proves $0 \leq \Phi \leq 1$. ∎

### 2.6.2 Parameters in NCL

In the proposed NCL method, there are two parameters $K$ and $\lambda$. $K$ $(0 < K < N-1)$ is the parameter in $K$-NN graph, which means each node has $K$ edges pointing from itself to its $K$ nearest neighbors. Following [11][19], we fix $K = 20$ in all our experiments. $\lambda$ $(0 \leq \lambda \leq 1)$ is the information threshold, which decides how much information is "enough" in learning cliques.

If $\lambda = 0$ and we delete all edges with zero weight. The clique of a node equals the connected graph that the node belongs to. Besides, cliques of nodes belongs to one connected graph are same. If the graph has only one connected graph, the collectiveness of the graph would be one. If $\lambda = 1$, the clique of a node only contains itself and the collectiveness of the graph equals zero.

## 2.7 Toy examples

To illustrate the performance and the property of the proposed method, we show two toy examples in this section. We compare the proposed NCL method with two methods which are the method in [11] and the method in [19]. To be concise, we name the method in [11] as $Z\_inv$ and name the method in [19] as $Z\_exp$.

### 2.7.1 Rectilinear graph

The rectilinear graph is the graph that has no circles and limited neighbors. As shown in Fig.4, the node in one directional rectilinear graph has no more than one neighbor and the node in bi-directional rectilinear graph has no more than two neighbors.

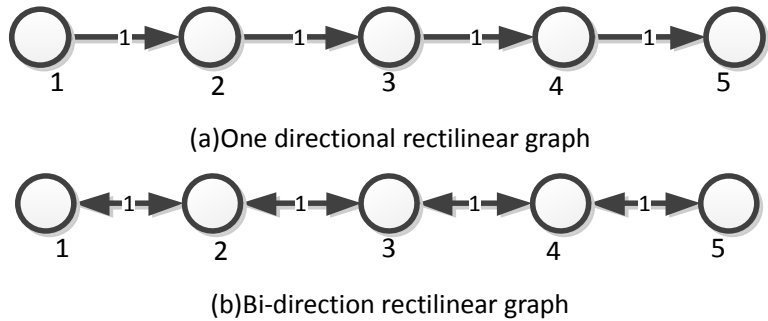

(a)One directional rectilinear graph

(b)Bi-direction rectilinear graph

Fig. 4. An example of rectilinear graph with five nodes.

In Fig.5, we compute the node collectiveness by different methods. For one directional rectilinear graph, NCL1 emphasizes the central nodes in the graph and NLC2 emphasizes the leading nodes in the graph. $Z\_inv$ and $Z\_exp$ emphasize the starting nodes in the graph. For bi-directional rectilinear graph, NCL1 and NCL2 treat one node just as important to the other nodes. $Z\_inv$ and $Z\_exp$ emphasize the central nodes in the graph.

**Property 5.** If all weights on edges equal one, we have $\Phi = 1/2$ (NCL1) and $\Phi = 3/4$ (NCL2) on one directional rectilinear graph, and $\Phi = 1$ (NCL1 and NCL2) on bi-directional rectilinear graph.

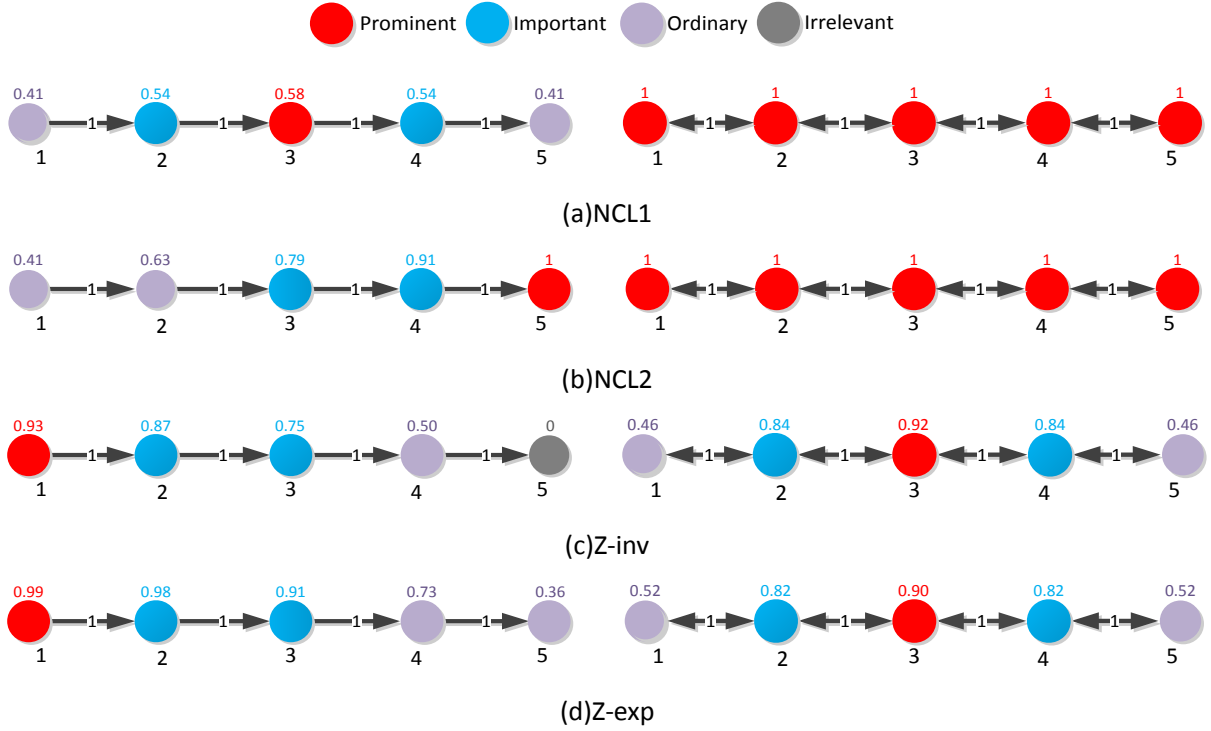

Fig. 5. Node collectiveness obtained by different methods. All weights on edges equal one. In Z-inv, $K=1$ for one directional rectilinear graph and $K=2$ for bi-directional rectilinear graph.

**Proof.** For one directional rectilinear graph, we know the node clique representation matrix $C \in R^{n\times n}$ is

$$C(i,j) = \begin{cases} 1, & if \ j \geq i \\ 0, & otherwise. \end{cases}$$

For NCL1, we have

$$J_1(i,j) = \begin{cases} \dfrac{n-i+1}{n-j+1}, & i > j \\ \dfrac{n-j+1}{n-i+1}, & i < j \end{cases}. \qquad J_2(i,j) = \begin{cases} \dfrac{j}{i}, & i > j \\ \dfrac{i}{j}, & i < j \end{cases}.$$

Then, it is easy to infer

$$\frac{\sum_{i=1,i\neq j}^{N}\sum_{j=1}^{N} J_1(i,j)}{N(N-1)} = \frac{\sum_{i=1,i\neq j}^{N}\sum_{j=1}^{N} J_2(i,j)}{N(N-1)} = \frac{1}{2}.$$

and

$$\Phi = \frac{\sum_{i=1}^{N} \dfrac{\sum_{j=1}^{N} Z(i,j)}{N-1}}{N} = \frac{1}{2}\left[\frac{\sum_{i=1,i\neq j}^{N}\sum_{j=1}^{N} J_1(i,j)}{N(N-1)} + \frac{\sum_{i=1,i\neq j}^{N}\sum_{j=1}^{N} J_2(i,j)}{N(N-1)}\right] = \frac{1}{2}.$$

For NCL2, we have

$$J_1'(i,j) = \begin{cases} 1, & i > j \\ \dfrac{n-j+1}{n-i+1}, & i < j \end{cases}. \qquad J_2'(i,j) = \begin{cases} \dfrac{j}{i}, & i > j \\ 1, & i < j \end{cases}.$$

Then, it is easy to infer

$$\frac{\sum_{i=1,i\neq j}^{N}\sum_{j=1}^{N} J_1'(i,j)}{N(N-1)} = \frac{\sum_{i=1,i\neq j}^{N}\sum_{j=1}^{N} J_2'(i,j)}{N(N-1)} = \frac{3}{4}.$$

and

$$\Phi = \frac{\sum_{i=1}^{N}\frac{\sum_{j=1}^{N}Z(i,j)}{N-1}}{N} = \frac{1}{2}\left[\frac{\sum_{i=1,i\neq j}^{N}\sum_{j=1}^{N}J_1'(i,j)}{N(N-1)} + \frac{\sum_{i=1,i\neq j}^{N}\sum_{j=1}^{N}J_2'(i,j)}{N(N-1)}\right] = \frac{3}{4}.$$

For bi-directional rectilinear graph, we know the node clique representation matrix $C \in R^{n \times n}$ is
$$C(i,j) = 1.$$
We have
$$J_1(i,j) = J_2(i,j) = J_1'(i,j) = J_2'(i,j) = 1.$$
Then, it is easy to infer $\Phi = 1$. ∎

### 2.7.1 Circle graph

The circle graph is a connected graph with only one circle, as shown in Fig.6.

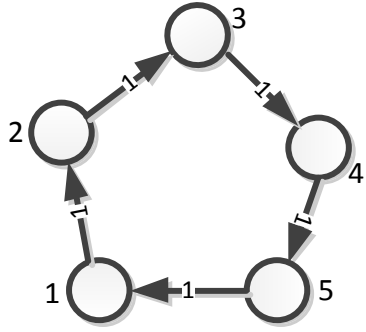

Fig. 6. An example of circle graph with five nodes.

**Property 6.** If all weights on edges equal one, we have $\Phi = 1$ (NCL1 and NCL2) on circle graph.
**Proof.** For circle graph, we know the node clique representation matrix $C \in R^{n \times n}$ is
$$C(i,j) = 1.$$
We have
$$J_1(i,j) = J_2(i,j) = J_1'(i,j) = J_2'(i,j) = 1.$$
Then, it is easy to infer $\Phi = 1$. ∎

If weights on edges are less than one, the graph collectiveness obtained by NCL will decrease as the growth of the number of nodes, as seen in Fig.7. It can be assumed that the circle graph without strongest connections will become looser when the graph becomes larger. However, graph collectiveness obtained by $Z\_inv$ and $Z\_exp$ do not change with different number of nodes.

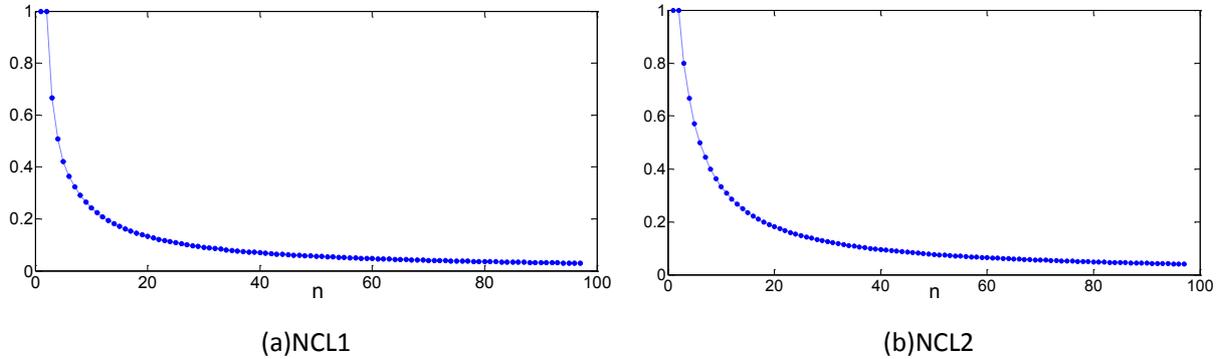

(a)NCL1  (b)NCL2
Fig. 7. Weights on edged equal 0.9 and $\lambda = 0.6$ for NCL methods.

# 3 Collectiveness experiment

In this section, we compare the proposed NCL method with two methods which are the $Z\_inv$ [11] and $Z\_exp$ [19].

## 3.1 SDP model

The Self-Driven Particle (SDP) model [13] is a famous model for studying collective motion and shows high similarity with various crowd systems in nature [14]. The ground-truth of collectiveness in SDP is the average normalized velocity $v = ||\frac{1}{N}\sum_{i=1}^{N}\frac{v_i}{||v_i||}||$, which was commonly used as a measure of collectiveness in existing works [11][20][21]. SDP model produce a system of moving particles that are driven with a constant speed [11]. SDP gradually turns into collective motion from disordered motion [11]. Each particle will update its direction of velocity to the average direction of the particles in its neighborhood at each frame [11]. The update of velocity direction $\theta$ [13] for every particles $i$ in SDP is

$$\theta_i(t+1) = <\theta_j(t)>_{j\in\mathcal{N}(i)} + \Delta\theta. \tag{12}$$

where $<\theta_j(t)>_{j\in\mathcal{N}(i)}$ denotes the average direction of velocities of particles within the neighborhood $\mathcal{N}(i)$ of $i$. $\Delta\theta$ is a random angle chosen with a uniform distribution within the interval $[-\eta\pi, \eta\pi]$, where $\eta$ tunes the noise level of alignment [13].

### 3.1.1 Neighborhood graph

Given $N$ moving particles in SDP, we can measure the similarity of particles by $K$-NN graph. For simplicity, we adopt the method in [11] to compute the $K$-NN graph for comparison. At time $t$, the weight value on edge between particle $i$ and particle $j$ are defined by [11]

$$w_t(i,j) = \begin{cases} max(\frac{v_i v_j^T}{||v_i||_2 ||v_j||_2}, 0), & if\ j \in \mathcal{N}(i) \\ 0, & otherwise \end{cases}. \tag{13}$$

where $v_i$ is the velocity vector of particle $i$.

### 3.1.2 Evaluation metrics of SDP

To evaluate the performance of different methods in SDP, we use the following three metrics:
1. Relevant Coefficient (**RC**). For each method, we compute the collectiveness of all frames and then compute the relevant coefficient between the measured collectiveness and the ground truth. We can get **higher** RC if the method is better.
2. Pairs Comparing Accuracy (**PCA**). For each method, we first compute the collectiveness of all frames. Then we compare all possible frame pairs by collectiveness. If there are $l$ frames, we have $\frac{l(l-1)}{2}$ pairs in in total. Based on the ground truth, the accuracy rate of judging right is defined as the Pairs Comparing Accuracy (PCA). We can get **higher** PCA if the method is better.
3. Sorting Difference (**SD**). For each method, we first compute the collectiveness of all frames. Then we sort the measured collectiveness and the ground truth in ascending or descending manner. The average absolute difference between the order number of the measured collectiveness and the order number of the ground truth is named as Sorting Difference (SD). We can get **lower** SD if the method is better.

The above three metrics evaluate a method from different perspectives.

### 3.1.3 Numerical analysis without noise

The parameters of SDP are fixed following [11][19]: $N = 400$, $K = 20$, size of ground $L = 7$, the absolute value of velocity $||v|| = 0.03$, the interaction radius $r = 1$ and $\eta = 0$. The parameters in Z_inv and Z_exp follow [11] and [19], respectively. We set $\lambda = 0.7$ in NCL.

We illustrate the performance of different methods (400 runs) in Tab.1. In each run, the SDP evolution ends if it exceeds 100 frames or the ground truth exceeds 0.95 for compute efficiency. Note that the SDP run might out of control every once in a while. NCL methods especially NCL_min methods perform well on three evaluation metrics. We also illustrate the performance of NCL with different ($0 \leq \lambda \leq 1$) (400 runs) in Tab.2. $\lambda$ varies from 0.1 to 0.99999999. Both NCL_avg and NCL_min are not sensitive to the parameter $\lambda$ even when $\lambda$ is close to one. Moreover, NCL_min is more robust to $\lambda$ than NCL_avg.

Tab.1 Performance of different method on SDP model. Bolded values in each row indicate best performance in RC, PCA and SD, respectively. Underlined values in each row indicate second best performance. Parameters: $\lambda = 0.7$ and $K = 20$.

|  | Z_inv | Z_exp | NCL1_avg | NCL2_avg | NCL1_min | NCL2_min |
|---|---|---|---|---|---|---|
| Relevant Coefficient (**RC**) | 0.850 | **0.943** | 0.942 | 0.933 | 0.895 | 0.923 |
| Pairs Comparing Accuracy (**PCA**) | 91.4% | 92.0% | 93.3% | 92.5% | 96.1% | **96.2%** |
| Sorting Difference (**SD**) | 4.765 | 4.551 | 3.783 | 4.315 | 2.384 | **2.342** |

Tab.2 Performance of NCL with different $\lambda$ on the SDP model. For each metric, bolded values and underlined values in each column indicate best and second best performance, respectively. Relevant Coefficient (RC): The higher, the better. Pairs Comparing Accuracy (PCA): The higher, the better. Sorting Difference (SD): The lower, the better. $\lambda$ varies from 0.1 to 0.99999999. Parameter: $K = 20$.

|  | $\lambda$ | 0.1 | 0.2 | 0.3 | 0.4 | 0.5 | 0.6 | 0.7 | 0.8 | 0.9 | 0.95 | 0.98 | 0.99 | 0.9999 | 0.999999 | 0.99999999 |
|---|---|---|---|---|---|---|---|---|---|---|---|---|---|---|---|---|
| **RC** | NCL1_avg | 0.644 | 0.730 | 0.801 | 0.870 | 0.891 | 0.932 | **0.942** | 0.934 | 0.906 | 0.886 | 0.852 | 0.840 | 0.816 | 0.793 | 0.653 |
|  | NCL2_avg | 0.591 | 0.667 | 0.737 | 0.814 | 0.847 | 0.904 | 0.933 | **0.942** | 0.920 | 0.897 | 0.866 | 0.853 | 0.818 | 0.787 | 0.654 |
|  | NCL1_min | **0.944** | 0.941 | 0.934 | 0.930 | 0.905 | 0.908 | 0.895 | 0.875 | 0.872 | 0.860 | 0.835 | 0.826 | 0.841 | 0.818 | 0.661 |
|  | NCL2_min | 0.933 | **0.943** | 0.945 | 0.951 | 0.929 | 0.936 | 0.923 | 0.901 | 0.890 | 0.880 | 0.856 | 0.846 | 0.837 | 0.807 | 0.661 |
| **PCA** | NCL1_avg | 75.6% | 80.3% | 84.2% | 86.8% | 88.9% | 91.8% | 93.3% | 94.5% | 95.8% | 96.0% | 95.5% | 95.3 | 92.3% | 89.0% | 77.6% |
|  | NCL2_avg | 75.1% | 79.5% | 83.0% | 85.3% | 87.4% | 90.7% | 92.5% | 94.1% | 95.7% | 96.0% | 95.7% | 95.6 | 92.5% | 89.0% | 77.9% |
|  | NCL1_min | 91.7% | 92.9% | 93.7% | 94.5% | 94.9% | 95.7% | 96.1% | 96.2% | 96.3% | 96.0% | 95.5% | 95.1 | 92.7% | 89.2% | 77.6% |
|  | NCL2_min | 91.3% | 92.7% | 93.6% | 94.5% | 94.9% | 95.8% | 96.2% | 96.4% | **96.6%** | 96.3% | 95.9% | 95.5 | 92.9% | 89.3% | 78.0% |
| **SD** | NCL1_avg | 7.518 | 6.823 | 6.294 | 5.817 | 5.330 | 4.362 | 3.783 | 3.309 | 2.553 | 2.429 | 2.574 | 2.701 | 3.991 | 5.113 | 8.399 |
|  | NCL2_avg | 7.698 | 7.266 | 6.972 | 6.696 | 6.272 | 5.068 | 4.315 | 3.571 | 2.692 | 2.458 | **2.525** | 2.585 | 3.883 | 5.106 | **8.365** |
|  | NCL1_min | **4.179** | 3.924 | 3.683 | 3.304 | 3.029 | 2.607 | 2.384 | 2.404 | 2.197 | 2.343 | 2.593 | 2.820 | 3.827 | 4.962 | 8.444 |
|  | NCL2_min | 4.528 | 4.124 | 3.836 | 3.414 | 3.146 | **2.589** | **2.342** | **2.340** | **2.088** | 2.214 | 2.431 | 2.611 | 3.715 | 4.958 | 8.369 |

### 3.1.4 Numerical analysis with noise

To evaluate different methods' performance on SDP model with noise data, we consider noise particles in SDP mode. In this section, we follow the parameter settings in 3.1.3: $N = 400$, $K = 20$, $L = 7$, $||v|| = 0.03$, $r = 1$ and $\eta = 0$. Thus, all particles including noise particles have the absolute value of velocity $||v|| = 0.03$. However, the noise particles have different motion model. At the beginning, the noise particles' spatial locations and velocity directions are randomly assigned. Then, the noise particles move towards a random direction at every frame. Note that the total number of particles $N$ equals 400. Thus, the more noise particles exist, the less normal particles exist. We use the ratio of noise particles to indicate the degree of noise. Note that the ground truth in noise case is the average normalized velocity of normal particles only. As seen in Fig.8, cases with different ratio of noise particles are shown. We evaluate different methods ($\lambda = 0.7$ for NCL) on the noise SDP model, and the ratio of noise particles varies from 0 to 0.8, as seen in Tab.3. NCL performs better than the compared methods on the noise SDP model.

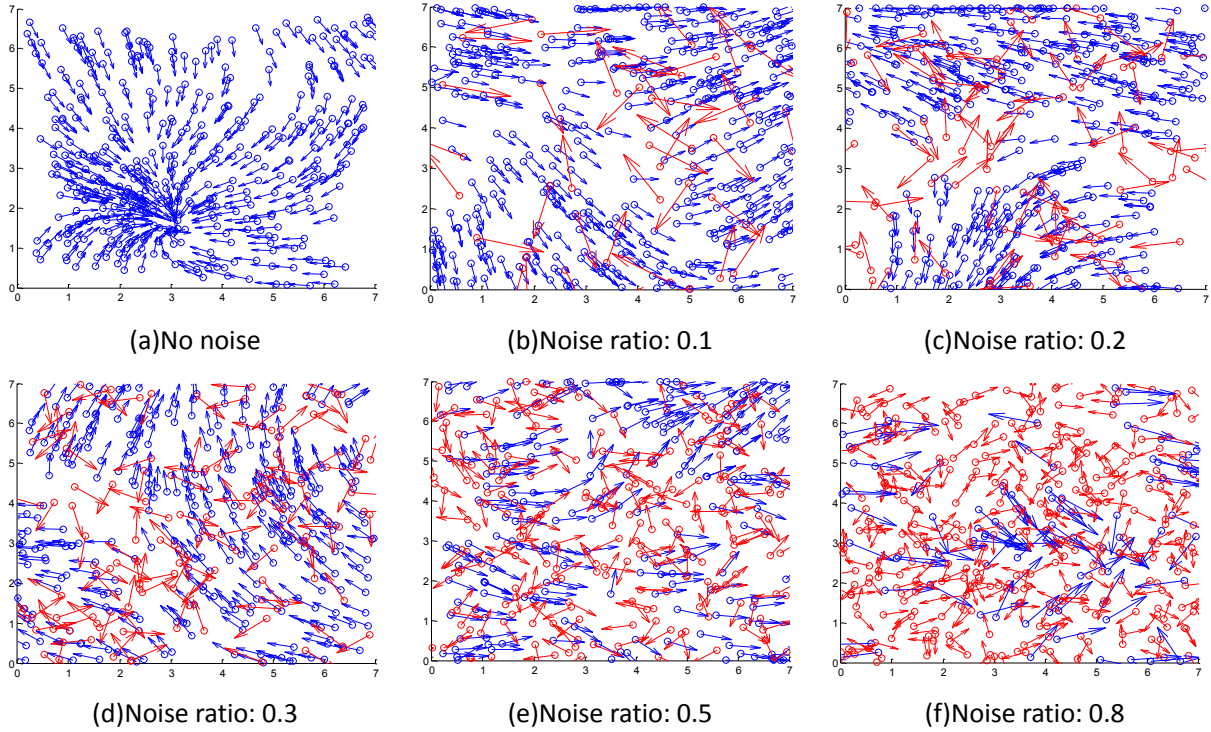

(a)No noise  (b)Noise ratio: 0.1  (c)Noise ratio: 0.2
(d)Noise ratio: 0.3  (e)Noise ratio: 0.5  (f)Noise ratio: 0.8

Fig. 8. Samples of SDP model with different ratio of noise particles.

Tab.3 Experiments on the noise SDP model. The ratio of noise particles varies from 0 to 0.8. For each metric, bolded values and underlined values in each column indicate best and second best performance, respectively. Parameters: $\lambda = 0.7$ and $K = 20$.

| The ratio of noise particles | | 0 | 0.1 | 0.2 | 0.3 | 0.4 | 0.5 | 0.6 | 0.7 | 0.8 |
|---|---|---|---|---|---|---|---|---|---|---|
| **Relevant Coefficient (RC)** | Z_inv | 0.850 | 0.865 | 0.857 | 0.827 | 0.799 | 0.709 | 0.571 | 0.354 | 0.137 |
| | Z_exp | **0.943** | 0.911 | 0.852 | 0.763 | 0.620 | 0.475 | 0.331 | 0.184 | 0.052 |
| | NCL1_avg | 0.942 | **0.945** | 0.936 | 0.915 | 0.894 | 0.827 | 0.725 | 0.535 | 0.263 |
| | NCL2_avg | 0.933 | 0.944 | **0.938** | **0.919** | **0.904** | **0.836** | **0.731** | **0.547** | **0.273** |
| | NCL1_min | 0.895 | 0.902 | 0.904 | 0.892 | 0.879 | 0.816 | 0.709 | 0.502 | 0.215 |
| | NCL2_min | 0.923 | 0.926 | 0.923 | 0.908 | 0.892 | 0.826 | 0.715 | 0.513 | 0.227 |
| **Pairs Comparing Accuracy (PCA)** | Z_inv | 91.4% | 85.0% | 82.3% | 79.8% | 77.2% | 72.6% | 67.4% | 61.4% | 54.4% |
| | Z_exp | 92.0% | 87.0% | 84.3% | 81.5% | 75.9% | 69.4% | 63.2% | 56.9% | 51.8% |
| | NCL1_avg | 93.3% | 92.1% | 91.0% | **88.7%** | **86.2%** | **81.1%** | 75.5% | 68.6% | 58.5% |
| | NCL2_avg | 92.5% | 91.1% | 90.1% | 88.1% | 86.0% | **81.1%** | **75.6%** | **69.0%** | **58.7%** |
| | NCL1_min | 96.1% | 93.1% | 90.7% | 87.9% | 85.5% | 80.4% | 74.7% | 67.3% | 56.9% |
| | NCL2_min | **96.2%** | **93.6%** | **91.4%** | 88.5% | 85.9% | 80.6% | 74.8% | 67.7% | 57.2% |
| **Sorting Difference (SD)** | Z_inv | 4.765 | 7.158 | 8.503 | 10.266 | 11.73 | 16.72 | 23.01 | 27.73 | 31.65 |
| | Z_exp | 4.551 | 6.198 | 7.414 | 9.344 | 12.01 | 18.08 | 25.09 | 29.63 | 32.49 |
| | NCL1_avg | 3.783 | 4.474 | 5.540 | 7.329 | 9.211 | 14.46 | **20.89** | **25.94** | **31.08** |
| | NCL2_avg | 4.315 | 5.023 | 5.977 | 7.692 | 9.444 | 14.74 | 21.33 | 26.11 | 31.23 |
| | NCL1_min | 2.384 | 3.911 | 5.277 | 7.198 | **8.931** | **14.22** | 20.94 | 26.46 | 31.52 |
| | NCL2_min | **2.342** | **3.702** | **5.018** | **7.053** | 8.948 | 14.37 | 21.30 | 26.48 | 31.33 |

### 3.1.5 Parameter $K$

There are two parameters which are $K$ and $\lambda$ in the proposed NCL method. The performance of the NCL method with different $\lambda$ is shown in Tab.2. In Tab.1-Tab.3, we follow the parameter setting of $K$ in [11][19] to evaluate the NCL method. In this section, we illustrate the performance of the NCL method with different value of $K$, as seen in Tab.4. It shows that NCL is robust to the parameter $K$.

Tab.4 Performance of NCL methods with different $K$. Parameter: $\lambda = 0.7$. For each metric, bolded values and underlined values in each row indicate best performance and second best performance, respectively.

|  | Z_inv | Z_exp | NCL1_avg | NCL2_avg | NCL1_min | NCL2_min |
|---|---|---|---|---|---|---|
| **RC,** $K=10$ | 0.798 | 0.872 | 0.920 | 0.889 | <u>0.933</u> | **0.945** |
| **RC,** $K=15$ | 0.832 | 0.925 | **0.945** | <u>0.934</u> | 0.903 | 0.932 |
| **RC,** $K=20$ | 0.850 | **0.943** | <u>0.942</u> | 0.933 | 0.895 | 0.923 |
| **RC,** $K=25$ | 0.871 | **0.947** | <u>0.943</u> | 0.940 | 0.878 | 0.908 |
| **RC,** $K=30$ | 0.888 | 0.942 | **0.944** | **0.944** | 0.883 | 0.909 |
| **PCA,** $K=10$ | 89.3% | 90.0% | 89.6% | 88.4% | **94.4%** | <u>94.3%</u> |
| **PCA,** $K=15$ | 90.7% | 91.3% | 92.1% | 91.5% | <u>95.8%</u> | **96.0%** |
| **PCA,** $K=20$ | 91.4% | 92.0% | 93.3% | 92.5% | <u>96.1%</u> | **96.2%** |
| **PCA,** $K=25$ | 92.3% | 92.8% | 94.1% | 93.6% | <u>96.1%</u> | **96.3%** |
| **PCA,** $K=30$ | 93.2% | 93.5% | 94.6% | 94.1% | <u>96.4%</u> | **96.6%** |
| **SD,** $K=10$ | 5.670 | 5.347 | 5.538 | 6.307 | **3.344** | <u>3.491</u> |
| **SD,** $K=15$ | 5.186 | 4.898 | 4.249 | 4.699 | <u>2.646</u> | **2.574** |
| **SD,** $K=20$ | 4.765 | 4.551 | 3.783 | 4.315 | <u>2.384</u> | **2.342** |
| **SD,** $K=25$ | 4.384 | 4.203 | 3.401 | 3.744 | <u>2.400</u> | **2.325** |
| **SD,** $K=30$ | 3.863 | 3.854 | 3.109 | 3.527 | <u>2.123</u> | **2.073** |

## 3.2 Crowd collectiveness

We adopt the collective motion database collecting by [11] to analysis the collectiveness of crowd scene. The collective motion database consists of 413 video clips from 62 crowded scenes. Following settings in [11], the generalized KLT (gKLT) tracker method [17] is used as the feature extraction method and $K$-nn graph is adopt as the affinity matrix $W$ (notice that $0 \leq W \leq 1$). In [11], the collectiveness of 413 video clips is estimated by 10 subjects as the human-labeled ground truth. They independently rate the level of collective motions in all clips from three options: low (0 score), medium (1 score), and high (2 score). Then the collectiveness scores of all clips range from 0 to 20.

For each video clip, its collectiveness is the average value of collectiveness of all frames. The collectiveness categories of clips can be defined by: ①**Scores** [19]. The collectiveness of a clip is defined as: Low collectiveness (214 clips): $0 \leq score \leq 5$. Medium collectiveness (105 clips): $5 < score < 15$. High collectiveness (94 clips): $15 \leq score \leq 20$. ②**Voting** [11]. The collectiveness of a clip is the majority voting of subjects' collectiveness rating.

To evaluate the performance of different collectiveness measure methods, we compute the AUC value of the ROC curve for binary classification of high and low, high and medium, and medium and low categories. ROC and AUC permit the numbers of different classes be unbalanced, and the AUC is the most common used quantitative metric for binary classification.

As seen in Tab. 5, we illustrate the performance of different methods on the Crowd database. NCL method performs well, especially on the binary classification of high-low categories and medium-low categories of videos. In NCL methods, the NLC2-avg method always performs well. The parameter $K$ is set to 20. Note that we ensure the legality of $W$ by programming $W = max(0, W)$ and $W = min(1, W)$ in the code. If we ignore the legality of $W$, we might obtain slightly different results.

Tab.5 Performance of different methods on the Crowd database. Parameter: $K = 20$. Bolded values and underlined values in each column indicate best performance and second best performance, respectively.

| AUC value of the ROC curve | | By scores | | | By voting | | |
|---|---|---|---|---|---|---|---|
| | | high-low | high-medium | medium-low | high-low | high-medium | medium-low |
| | Z_inv | 0.950 | <u>0.852</u> | 0.770 | 0.939 | **0.827** | 0.801 |
| | Z_exp | 0.936 | 0.841 | 0.791 | 0.923 | <u>0.816</u> | 0.822 |
| $\lambda = 0.7$ | NCL1_avg | 0.991 | 0.847 | 0.849 | <u>0.988</u> | 0.815 | 0.874 |
| | NCL2_avg | **0.993** | **0.860** | <u>0.859</u> | **0.989** | **0.827** | <u>0.884</u> |
| | NCL1_min | 0.983 | 0.804 | 0.850 | 0.979 | 0.771 | 0.875 |
| | NCL2_min | 0.985 | 0.814 | <u>0.859</u> | 0.981 | 0.780 | <u>0.884</u> |
| $\lambda = 0.8$ | NCL1_avg | 0.991 | 0.842 | 0.850 | 0.987 | 0.810 | 0.875 |
| | NCL2_avg | **0.993** | 0.849 | **0.861** | **0.989** | 0.814 | **0.886** |
| | NCL1_min | 0.979 | 0.799 | 0.840 | 0.975 | 0.763 | 0.865 |
| | NCL2_min | 0.985 | 0.814 | 0.857 | 0.981 | 0.777 | 0.882 |
| $\lambda = 0.9$ | NCL1_avg | 0.987 | 0.832 | 0.843 | 0.983 | 0.799 | 0.868 |
| | NCL2_avg | <u>0.992</u> | 0.843 | 0.856 | <u>0.988</u> | 0.809 | 0.881 |
| | NCL1_min | 0.966 | 0.779 | 0.824 | 0.961 | 0.740 | 0.849 |
| | NCL2_min | 0.981 | 0.807 | 0.847 | 0.977 | 0.768 | 0.872 |

We analysis the collective motion in clip by the threshold clustering method, which is used in [11] and [19]. We can get the clusters of collective motion patterns as the connected components by setting a clustering threshold and thresholding the values on $Z$. Figure 9 illustrates the performance of extracting collective motions. In fact, the clustering threshold is very important in collective motion extracting. One can adjust the clustering threshold for each clip to extract the most reasonable collective motions. However, the best clustering threshold might be different across different scenes. In Fig.10, we illustrate examples of extracting collective motions by different clustering threshold. It can be find out that the proposed NCL2_avg is robust to the clustering threshold parameter and performs well. It is also reasonable to assume that clips with low collectiveness have more trivial clusters than clips with high collectiveness.

Though the proposed NCL is robust to parameters, it is an important job to find suitable parameters. For selecting parameters in NCL, we can use the finite grid method [18].

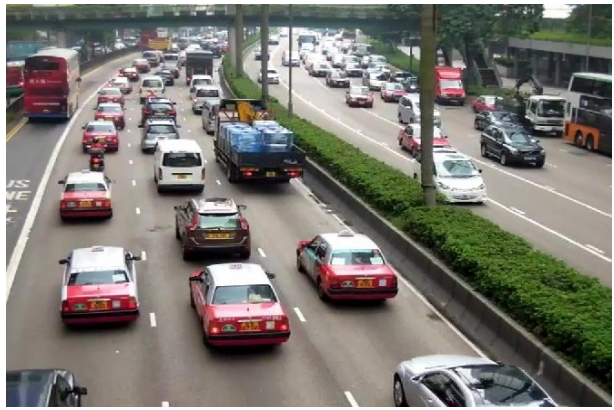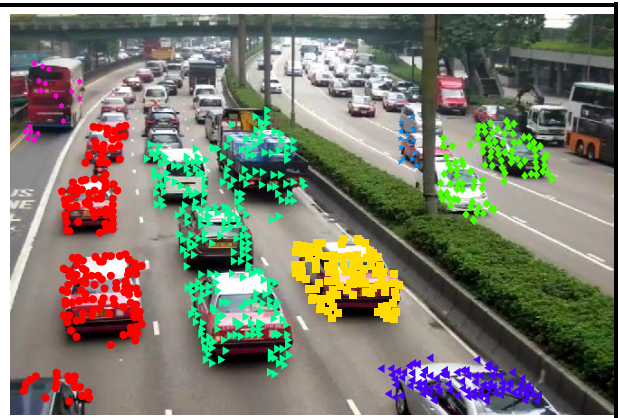

Clip:'4wanchaitraffic5';score:19;Voting:high.   Z_inv, 7 clusters

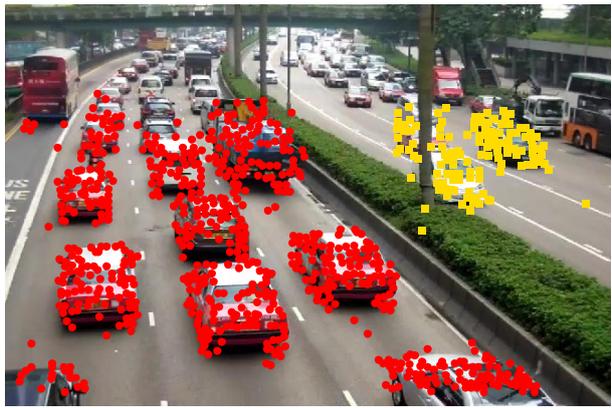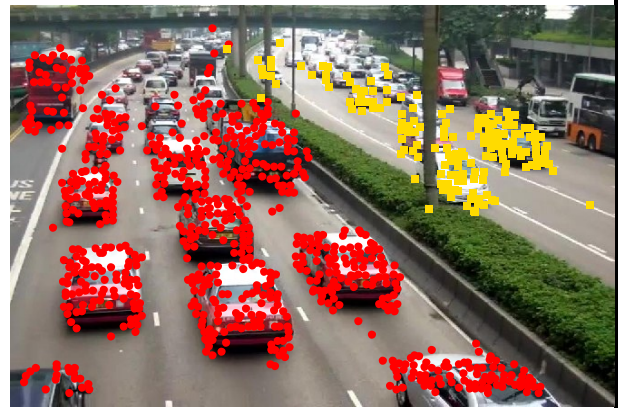

Z_exp, 2 clusters   NCL2_avg, 2 clusters

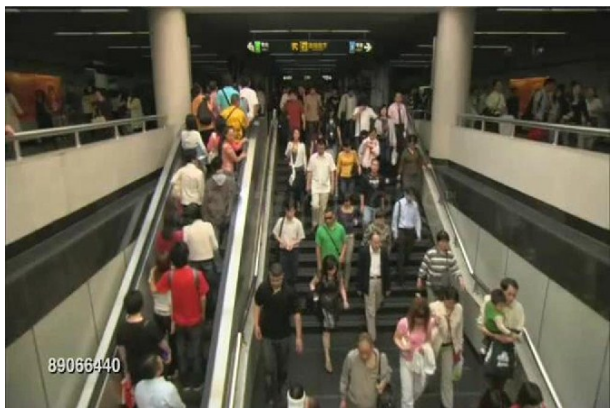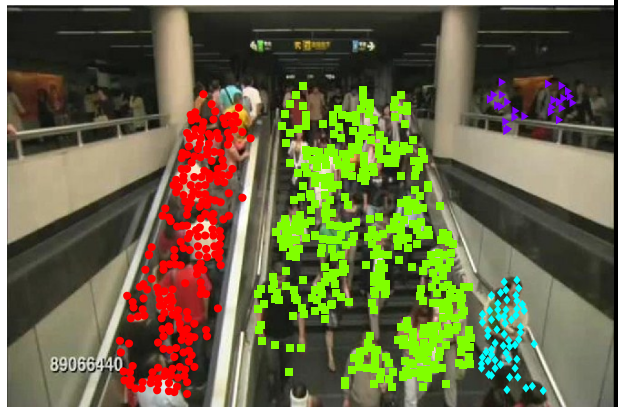

Clip:'laddergetty1';score:14; Voting:medium.   Z_inv, 4 clusters

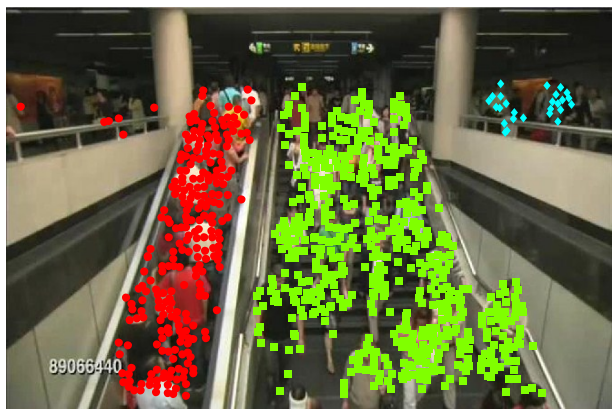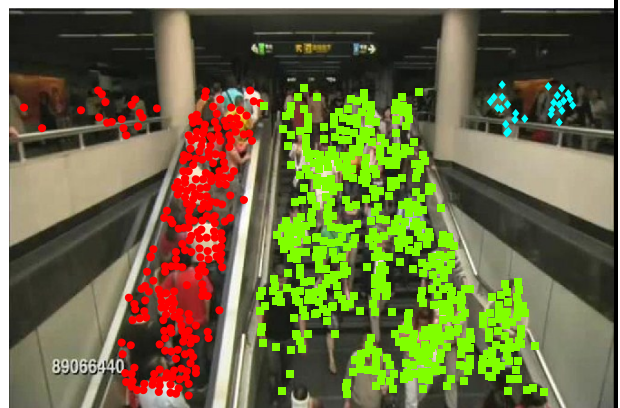

Z_exp, 3 clusters   NCL2_avg, 3 clusters

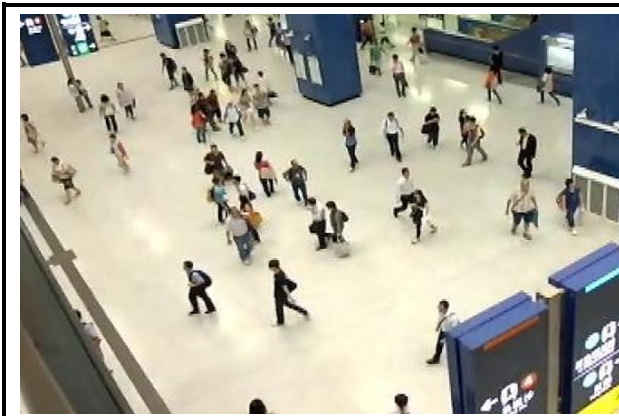
Clip:'1dawei2';score:1;Voting:low

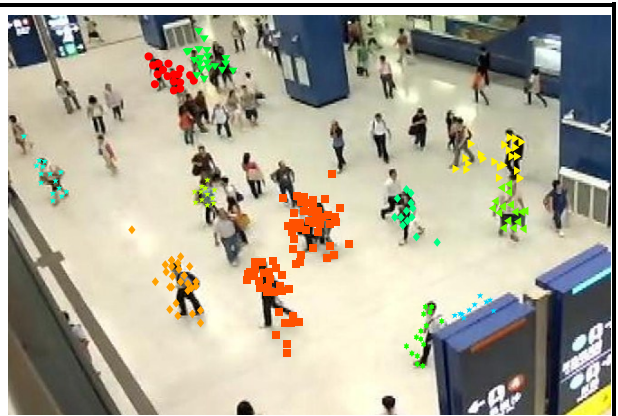
Z_inv, 11 clusters

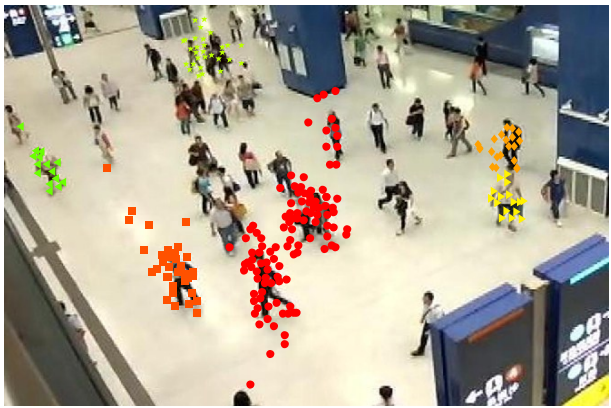
Z_exp, 6 clusters

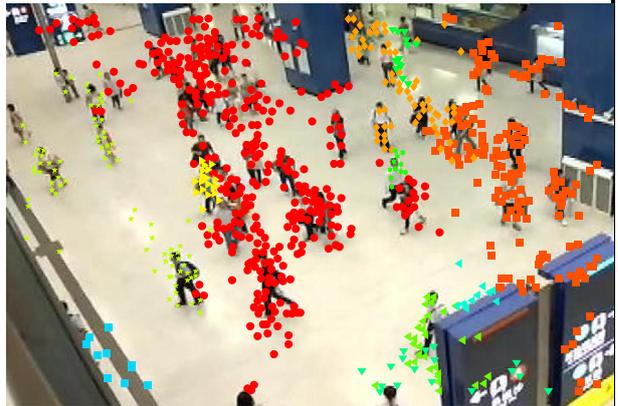
NCL2_avg, 11 clusters

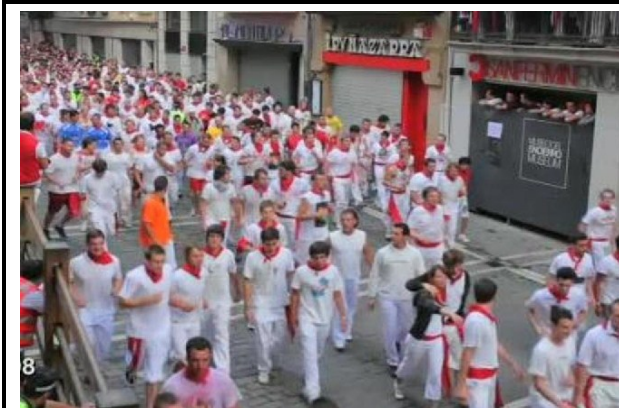
Clip:'niurunning1';score:18;Voting:high

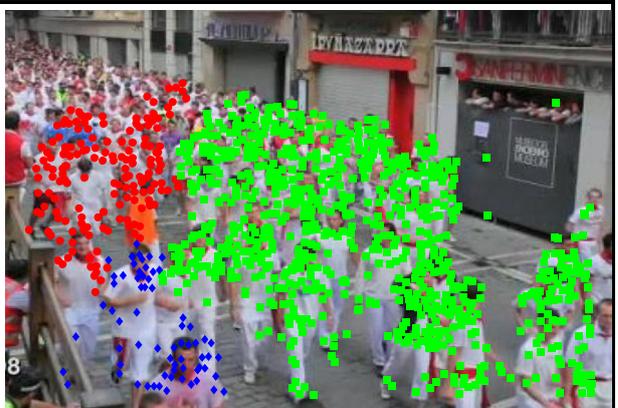
Z_inv, 3 clusters

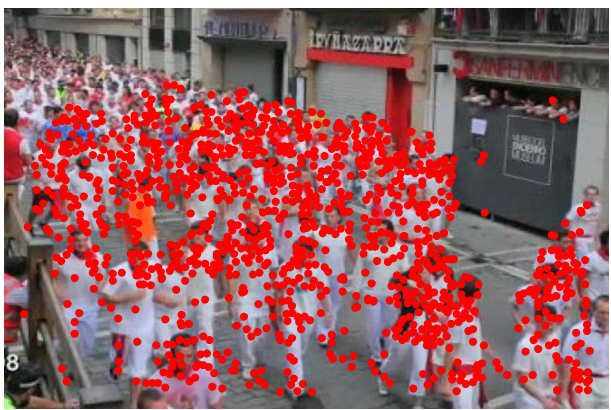
Z_exp, 1 clusters

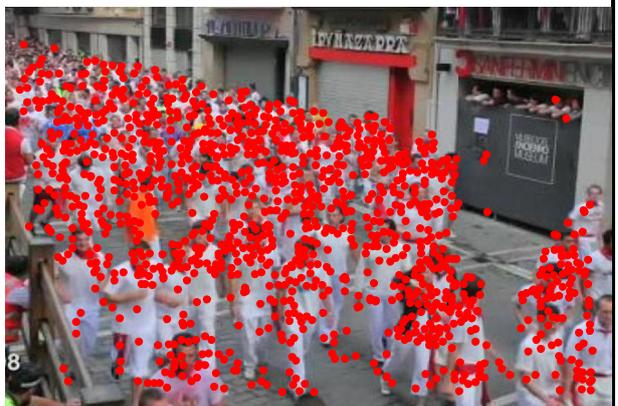
NCL2_avg, 1 clusters

Fig. 9. Performance illustrations of extracting collective motions (frame 10). Different collective motions are illustrated by different colors. The clustering threshold for Z_inv, Z_exp and NCL2_avg is set to 0.03, $10^{-5}$ and 0.4, respectively. Parameter: $K = 20$ for all methods. $\lambda = 0.7$ for NCL2_avg.

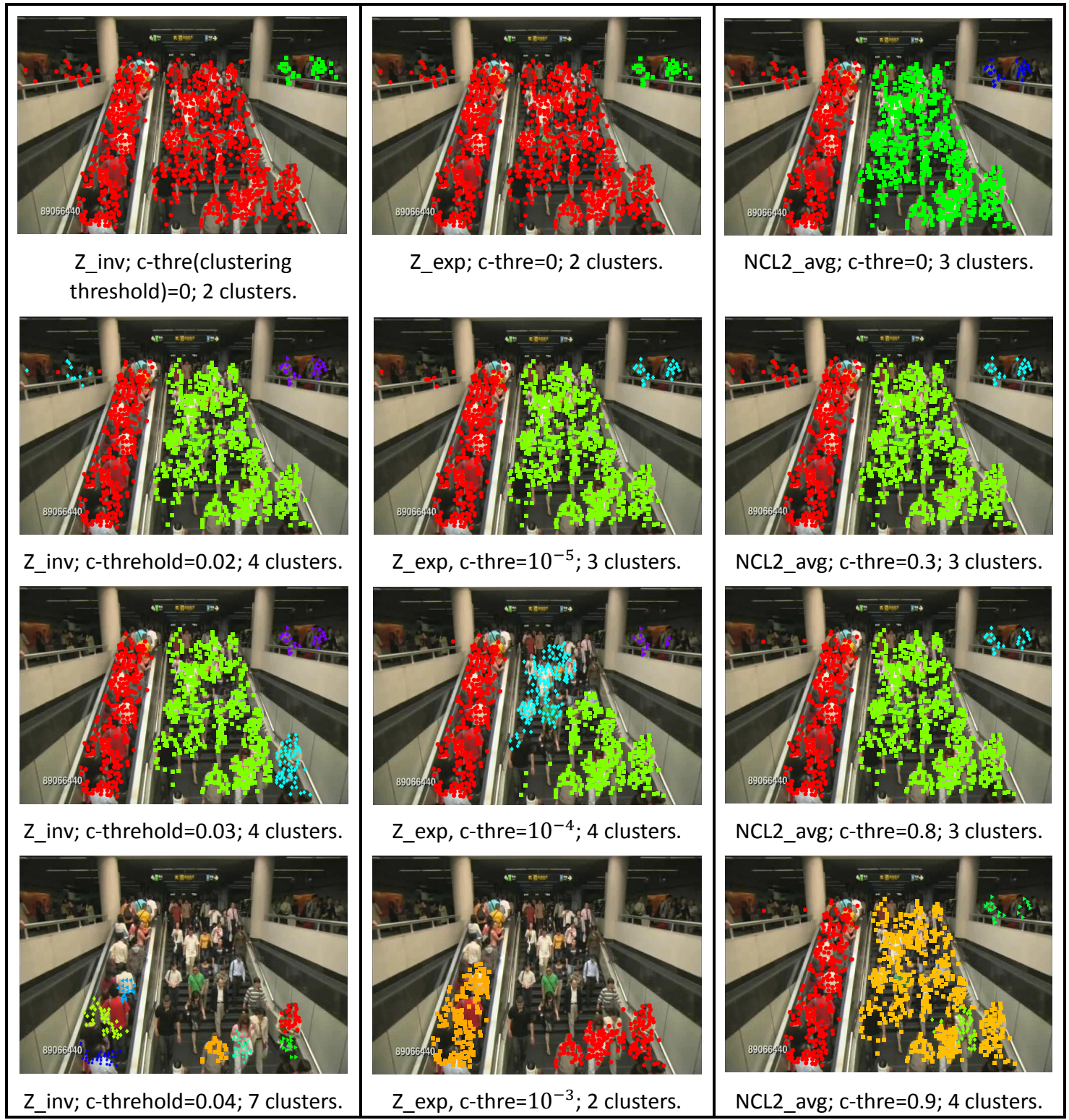

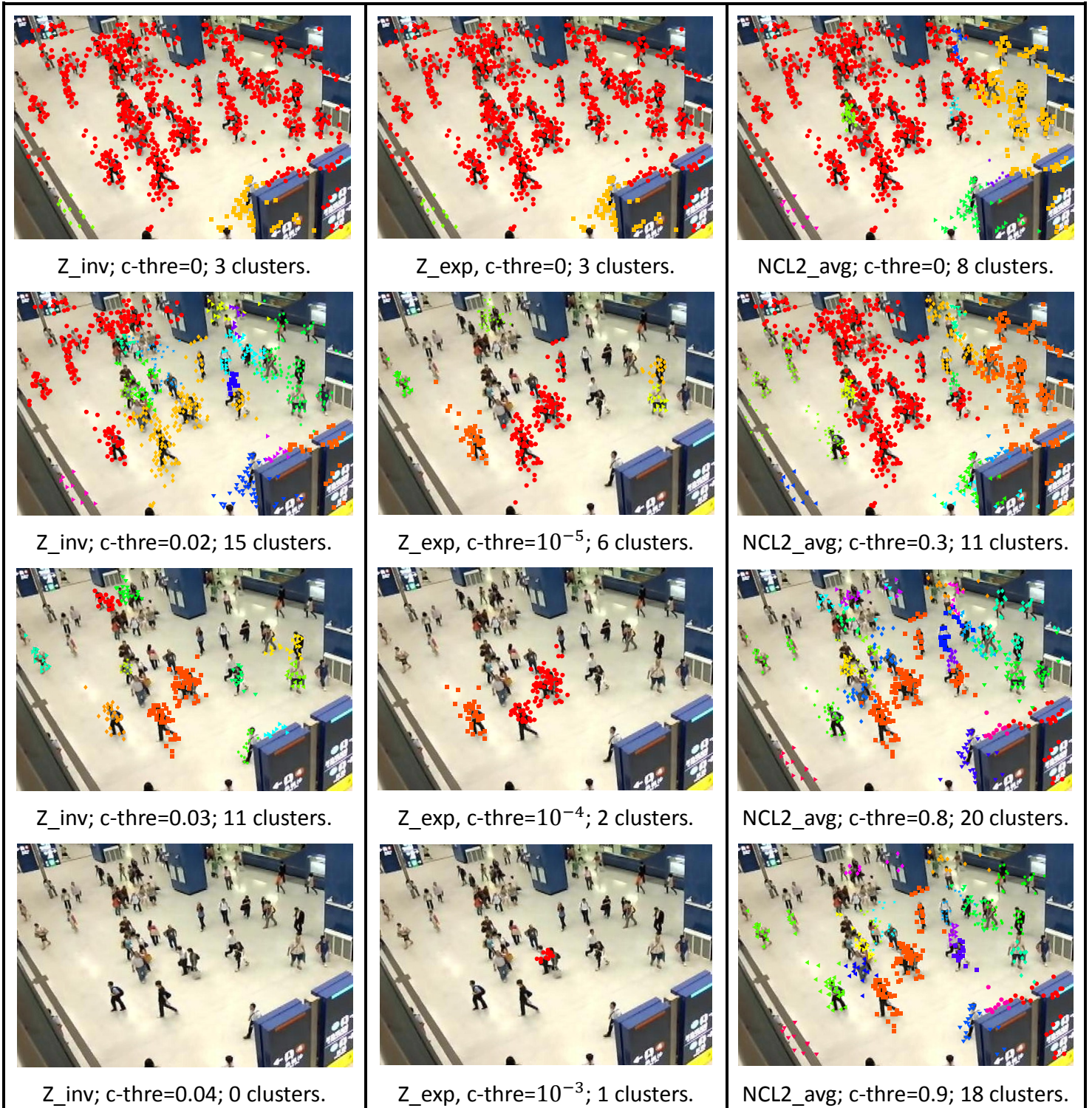

Fig. 10 Performance of collective motion extraction with different clustering threshold. The clustering threshold is denoted as c-thre for short.

## 4 Conclusion

In this paper, we proposed a node clique learning method named NCL for representing nodes in graph. Node clique of a node reflects the node's influence on other nodes. We can compute the nodes' coherence by comparing their node cliques. Several fine properties of the NCL have been shown.

Experiment on SDP model and crowd scene database show the good performance of the proposed method in measuring collectiveness. The shortage of the proposed method is the problem computational efficiency because we need to compute the cliques of each node in graph. To solve this problem, parallel computing can be considered in NCL algorithm to improve computational efficiency. In NCL algorithm, we update a ready node's privileged state after information computing right away. Then one can update the ready nodes' privileged state after all ready nodes' information is computed to improve the computational efficiency. The relationship between the number of collective motions and human cognition can be considered to improve the proposed method. In future work, we will study the online version of the proposed method. Besides, we will compute the collectiveness of various types of crowd systems, including more real-task crowd scenes.

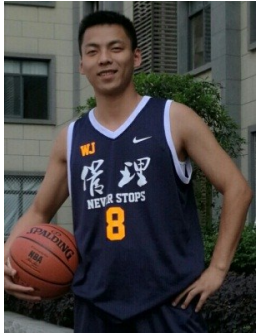

**Weiya Ren** received his B.S. and Ph.D. degree in automotive engineering and system engineering from National University of Defense Technology, Changsha, China, in 2010 and 2015, respectively. He is currently working at Department of Management Science and Engineering, Officers College of Chinese Armed Police Force, Chengdu, China. His current research interests include machine learning, computer vision, the internet of things and complex network.